\documentclass{naturep}

\usepackage{amssymb}
\usepackage{amsmath}
\usepackage{graphicx}
\usepackage{algorithmicx}
\usepackage{algorithm}
\usepackage{subfigure}
\usepackage{subcaption}
\usepackage[noend]{algpseudocode}
\usepackage{bbm}
\usepackage{lineno}
\usepackage[margin=0.6in]{geometry}
\usepackage{ragged2e}
\usepackage{setspace}
\usepackage{longtable}
\usepackage{hyperref}
\usepackage{anyfontsize}
\usepackage{setspace}
\usepackage{float}
\usepackage{booktabs}
\usepackage{multirow}
\usepackage{xcolor}
\usepackage{blindtext}
\usepackage{tabularx}
\usepackage{caption}
\usepackage{array}
\usepackage{bbding}
\usepackage{pifont}
\usepackage{fontawesome}
\usepackage{lineno}
\usepackage{graphicx}

\usepackage[normalem]{ulem}

\usepackage{subcaption}

\makeatletter
\let\saved@includegraphics\includegraphics
\AtBeginDocument{\let\includegraphics\saved@includegraphics}
\renewenvironment*{figure}{\@float{figure}}{\endfloat}
\makeatother

\usepackage{enumitem, kantlipsum}

\title{OphMAE: Bridging Volumetric and Planar Imaging with a Foundation Model for Adaptive Ophthalmological Diagnosis}

\begin{document}

\maketitle
\begin{spacing}{1.8}
\vspace{-10mm}
\noindent Tienyu Chang$^{1}$$^{\boldsymbol{\ddag}}$, Zhen Chen$^{2,3}$$^{\boldsymbol{\ddag}}$,  Renjie Liang$^{5}$, Jinyu Ding$^{2}$, Jie Xu$^{5}$, Sunu Mathew$^{6,7}$, Amir Reza Hajrasouliha$^{7}$, Andrew J. Saykin$^{6, 8}$, Ruogu Fang$^{4}$*, Yu Huang$^{1}$*, Jiang Bian$^{1}$*, Qingyu Chen$^{2}$* 
\end{spacing}
\vspace{-10mm}
\begin{spacing}{1.4}
\begin{affiliations}
\item Department of Biostatistics and Health Data Science, Indiana University, Indianapolis, IN
\item Department of Biomedical Informatics and Data Science, Yale University, New Haven, CT
\item Department of Data Science and Artificial Intelligence, The Hong Kong Polytechnic University, Hong Kong
\item Department of Biomedical Engineering, University of Florida, Gainesville, FL
\item Department of Health Outcomes and Biomedical Informatics, University of Florida, Gainesville, FL
\item Radiology \& Imaging Sciences, Indiana University, Indianapolis, IN
\item Ophthalmology, Indiana University, Indianapolis, IN
\item Center for Neuroimaging and Indiana Alzheimer's Disease Research Center, Indiana University, Indianapolis, IN

$^{\boldsymbol{\ddag}}$Contributed Equally\\
\textbf{*Corresponding authors}
\end{affiliations}
\end{spacing}

\vspace{-5mm}
\begin{spacing}{1.0}
\section{Abstract} 
The advent of foundation models has heralded a new era in medical artificial intelligence (AI), enabling the extraction of generalizable representations from large-scale unlabeled datasets. However, current ophthalmic AI paradigms are predominantly constrained to single-modality inference, thereby creating a dissonance with clinical practice where diagnosis relies on the synthesis of complementary imaging modalities. Furthermore, the deployment of high-performance AI in resource-limited settings is frequently impeded by the unavailability of advanced three-dimensional imaging hardware. Here, we present the Ophthalmic multimodal Masked Autoencoder (OphMAE), a multi-imaging foundation model engineered to synergize the volumetric depth of 3D Optical Coherence Tomography (OCT) with the planar context of 2D en face OCT. By implementing a novel cross-modal fusion architecture and a unique adaptive inference mechanism, OphMAE was pre-trained on a massive dataset with of 183,875 paired OCT images derived from 32,765 patients. In a rigorous benchmark encompassing 17 diverse diagnostic tasks with 48,340 paired OCT images from 8,191 patients, the model demonstrated state-of-the-art performance, achieving an Area Under the Curve (AUC) of 96.9\% for Age-related Macular Degeneration (AMD) and 97.2\% for Diabetic Macular Edema (DME), consistently surpassing existing single-modal and multimodal foundation models. Crucially, OphMAE exhibits robust engineering adaptability: it maintains high diagnostic accuracy, such as 93.7\% AUC for AMD, even when restricted to single-modality 2D inputs, and demonstrates exceptional data efficiency by retaining 95.7\% AUC with as few as 500 labeled samples. This work establishes a scalable and adaptable framework for ophthalmic AI, ensuring robust performance across different tasks.

\end{spacing}


\newpage


\section{Introduction}

Ophthalmic diseases represent a major global health burden that causes irreversible vision loss, which emphasizes the necessity of early screening and the need for accurate, accessible diagnostic tools \cite{burton2021lancet,bourne2021trends,flaxman2017global}.  In clinical environments, ophthalmologists utilize multiple imaging tools to diagnose eye diseases, such as Optical Coherence Tomography (OCT), Optical Coherence Tomography Angiography (OCTA), Fluorescein Angiography, Fundus Photography, Fundus Autofluorescence (FAF), etc. \cite{amd_guide,dr_guide,glaucoma_guide} Among them, OCT has become gold standard of initial diagnosis tool for many eye diseases \cite{amd_guide,dr_guide,amd2}, e.g., age-related macular degeneration (AMD) and diabetic macular edema (DME), because it offers high-resolution visualization of retinal microstructure and supports precise evaluation retina pathological changes \cite{oct_adv1,oct_adv2}. Yet, the growth in imaging use has outpaced the ophthalmologist capacity, motivating automated methods that can support screening \cite{garvin2009automated}. In parallel, Artificial Intelligence (AI) has achieved diagnostic performance comparable to healthcare professionals across various medical imaging tasks \cite{ai_adv1,ai_adv2,ai_adv3}. 


\noindent In ophthalmic practice, the integration of Artificial Intelligence (AI) has also shown the potential for improving diagnostic accuracy and accessibility \cite{ting2019artificial,abramoff2018pivotal,gulshan2016development,de2018clinically}. Deep learning architectures have shown remarkable success in automated analysis of retinal imaging, with early studies demonstrating human-level performance in diabetic retinopathy screening from fundus photographs \cite{gulshan2016development} and subsequent advances enabling the detection of multiple retinal pathologies from optical coherence tomography (OCT) scans. \cite{de2018clinically} The field has recently witnessed the emergence of foundation models \cite{he2022masked}, \textit{i.e.}, large-scale neural networks pre-trained through self-supervised learning on extensive unlabeled datasets. Ophthalmic foundation models such as RETFound \cite{zhou2023foundation} and VisionFM \cite{qiu2024development} have demonstrated the capacity to learn generalizable ophthalmic representations that can be efficiently adapted to diverse diagnostic tasks through minimal fine-tuning. Additionally, OCTCube \cite{octcubem} further extends the input modality in 3D and demonstrates the strength of 3D models through detailed validation across cohorts and devices. These advances in the ophthalmic foundation model represent a paradigm shift from task-specific model development toward universal, adaptable AI systems for disease screening.


\noindent Despite these advances, existing approaches predominantly rely on single-modality imaging analysis, constraining their diagnostic scope relative to clinical practice standards. By processing these modalities in isolation, RETFound \cite{zhou2023foundation} is restricted to either 2D fundus images or OCT independently, thereby overlooking the complementary information embedded in paired data. While recent multimodal pretrained models like VisionFM \cite{qiu2024development} attempt to bridge this gap, they often lack specific engineering optimizations for the paired nature of multi-modality OCT data. 
Similarly, OCTCube \cite{octcubem} explores the integration of additional modalities into a 3D OCT framework; however, it does not explicitly address alignment between OCT volumes and en face images, nor does it provide a systematic evaluation comparing single-modality and multimodal performance. 

In routine ophthalmologic care, clinicians synthesize information across multiple imaging modalities to achieve comprehensive disease assessment, as each modality provides distinct yet complementary insights into ocular anatomy and pathophysiology \cite{schmidt2014guidelines}. For example, while OCT excels at revealing retinal micro-structural changes and detecting intraretinal or subretinal fluid, neovascular complexes characteristic of wet AMD often remain poorly visualized on structural OCT \cite{spaide2015retinal,freund2016retinal}. In contrast, en face OCT provides complementary, layer-resolved planar visualization that can enhance the assessment of spatial patterns (e.g., lesion distribution and subtle changes that may be less apparent on individual B-scans) \cite{feo2025face}. Importantly, en face OCT is acquired or generated alongside 3D B-scan OCT in clinical environments, and prior studies have demonstrated the diagnostic value of integrating en face views with other modalities across multiple ophthalmic conditions. \cite{enface_app1,enface_app2,enface_app3} This multi-modal integration is essential for accurate diagnosis, treatment planning, and monitoring of therapeutic response. Moreover, clinical settings frequently present constraints where only a subset of modalities are available due to equipment limitations, patient tolerance, or imaging quality issues, necessitating AI systems capable of robust performance across variable input configurations.

\noindent Despite the promise of the multimodal ophthalmic foundation model, a major challenge is how we learn the relation between modalities. 3D OCT volume and 2D en face OCT present a dimension asymmetry that standard self-supervised learning methods are not designed to handle, requiring novel modalities fusion and training strategies that simultaneously respect both 3D and 2D planar structures. Furthermore, cross-modal alignment poses a distinct challenge. Unlike image-text pairs, where contrastive objectives are well-motivated by semantic equivalence, OCT and en face OCT carry complementary and asymmetric information, making naive alignment objectives insufficient and potentially destabilizing to model training. Failure to resolve this challenge risks the model collapsing on one modality, suppressing the complementary contributions, and damaging model performance in downstream tasks. 

\begin{figure}[H]
  \centering
  \includegraphics[width=0.85\linewidth]{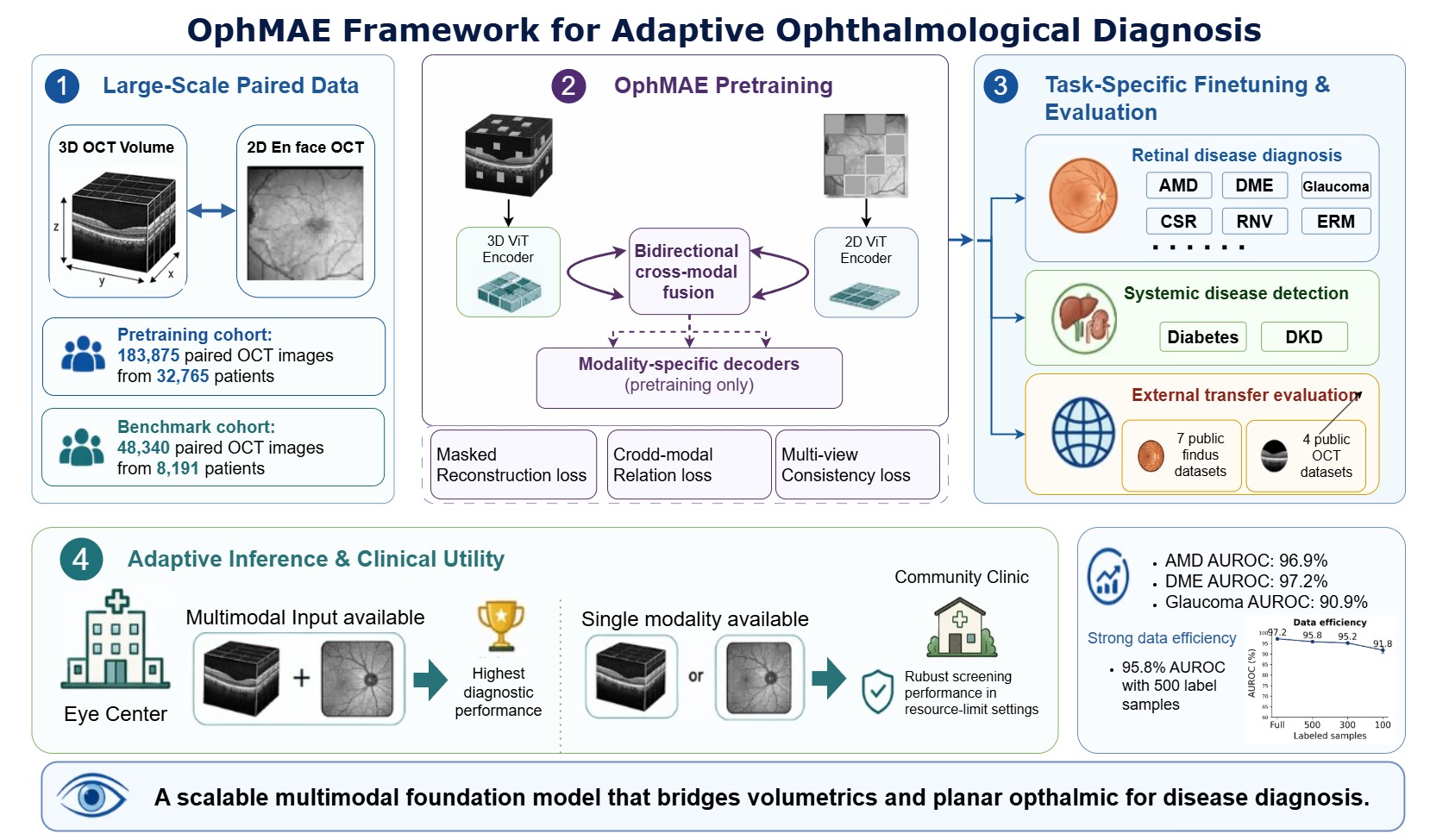}
  \caption{\textbf{Overview of the proposed OphMAE framework.} 
  \textbf{1)} Large-scale paired ophthalmic imaging data were curated from the University of Florida (UF) Health clinical data repository, comprising paired 3D OCT volumes and 2D en face OCT images. The dataset was divided into 183,875 paired images from 32,765 patients for self-supervised pretraining and 48,340 paired images from 8,191 patients for downstream benchmarking. 
  \textbf{2)} OphMAE pretraining integrates modality-specific 3D OCT and 2D en face OCT encoders through bidirectional cross-modal fusion, followed by modality-specific decoders used only during pretraining. The model is optimized using masked reconstruction loss, cross-modal relation loss, and multi-view consistency loss. 
  \textbf{3)} The pretrained model is fine-tuned and evaluated across retinal disease diagnosis, systemic disease detection, and external transfer benchmarks using public fundus and OCT datasets. 
  \textbf{4)} OphMAE supports adaptive inference in clinically heterogeneous settings, enabling multimodal inference when both imaging modalities are available and single-modality inference when only one modality is accessible. 
  \textbf{5)} The framework demonstrates strong diagnostic performance and data efficiency, maintaining robust AUROC with limited labeled samples.}
  \label{fig:model_overview}
\end{figure}

\noindent To address the fundamental limitation, we developed OphMAE (Figure \ref{fig:model_overview}), a foundation model that integrates volumetric OCT and en face OCT for enhanced diagnosis of retinal and systemic diseases through a unified framework. Our approach extends the masked autoencoder framework \cite{he2022masked} through novel cross-modal fusion architectures that enable effective multi-imaging representation learning. Additionally, we introduce innovative pre-training objectives, including cross-modal relation masking and multi-space semantic similarity consistency that harness the rich supervisory signals present in paired, unlabeled OCT-en face OCT datasets. The resulting model incorporates adaptive inference mechanisms that maintain high performance whether multimodal or single-modal inputs are available, addressing the heterogeneity of clinical imaging workflows. Through comprehensive validation on large-scale datasets from the University of Florida (48,340 paired OCT scans from 8,191 patients), our model demonstrates superior diagnostic performance compared to existing foundation models across multiple ophthalmic conditions. Additionally, we transfer our OphMAE and evaluate it on 7 public fundus and 4 public OCT datasets. This work establishes a new framework for clinically-relevant multi-modality imaging AI in ophthalmology, providing a scalable framework for improving diagnostic capabilities while maintaining the flexibility essential for real-world clinical deployment.

\section{Results}

\subsection{Study design}

\noindent Current ophthalmic foundation models are developed for a single imaging modality, even though routine retinal assessment often relies on complementary information from both volumetric OCT and en face OCT. OCT volumes provide retinal structural information that is essential for identifying retinal pathology, whereas en face OCT captures global spatial context and surface patterns that may facilitate localization and interpretation. This gap between model design and real-world clinical practice motivated the development of OphMAE, a foundation model designed to learn unified multimodal representations from paired ophthalmic images while remaining applicable when only one modality is available at inference.

\noindent We utilized a large-scale paired dataset of 3D OCT volumes and corresponding 2D en face OCT images from the University of Florida (UF) Health clinical data repository, where structured electronic health record (EHR) information has been harmonized into the Observational Medical Outcomes Partnership (OMOP) Common Data Model (CDM). We included patients who underwent ophthalmology visits or eye-related procedures involving retinal imaging between January 1, 2012, and January 1, 2024. All OCT volumes and en face OCT images were acquired using Heidelberg Spectralis systems. This paired acquisition enabled anatomically aligned multimodal pretraining, allowing OphMAE to learn both modality-specific features and cross-modal correspondences from unlabeled clinical data.


\noindent To systematically evaluate OphMAE for ophthalmic diagnosis, we established a comprehensive benchmark covering multiple diagnostic tasks across common and sight-threatening retinal diseases. The benchmark was designed to evaluate model performance in diverse clinical settings, including multimodal inference with both OCT and en face OCT, single-modality inference with only one available modality, and public dataset generalization. The training and internal validation dataset was collected from the University of Florida Health Eye Institute and consisted of paired three-dimensional optical coherence tomography (3D OCT) volumes and two-dimensional en face OCT images.

\noindent We compared our multimodal foundation model against several state-of-the-art ( models, including RETFound, a single-modality pretrained foundation model, VisionFM, a 2D visual foundation model developed for diverse modalities pretraining and single modality finetuning for downstream tasks, and OCTCube, a foundation model that utilizes 3D OCT volumes pretraining. These SOTA methods are common foundation models in ophthalmic and related disease diagnosis.

\subsection{Construction of large-scale paired multi-modal benchmark}

\noindent To systematically evaluate the efficacy of multi-imaging representation learning, we curated a large-scale pretraining dataset from the University of Florida Health Eye Institute using 80\% of the patient cohort, totaling 32,765 patients. Unlike prior datasets that rely on loosely matched temporal associations between OCT or en face photography, our dataset consists of strictly paired 3D OCT and 2D En face OCT images acquired simultaneously, ensuring precise anatomical alignment. The remaining cohort was reserved for downstream benchmarking, while the public datasets of fundus and 3D OCT data served as the external validation set to assess cross-institutional generalizability. The resulting UF benchmark spans 17 downstream diagnostic tasks across three settings: binary ophthalmic disease classification tasks, including Age-related macular degeneration (AMD), Cataract, Central retinal Vein occlusion/Central retinal artery occlusion (CRVO/CRAO), Central serous retinopathy (CSR), Diabetic macular edema (DME), Diabetic retinopathy (DR), Drusen, Epiretinal membrane (ERM), Macular hole (MH), Posterior vitreous detachment (PVD), and Retinal neovascularization (RNV); multi-class ophthalmic disease classification tasks, including DR, DME, and Glaucoma disease stages grading; and detection of systemic disease, such as Diabetes and Diabetic Kidney Disease (DKD). Detailed data processing and description are in the Supplemental material.
 
\noindent All models were trained and fine-tuned under identical experimental settings, including the same optimizer, learning rate, and number of training iterations. Performance was assessed using area under the receiver operating characteristic curve (AUROC), area under the precision-recall curve (AUPRC), accuracy, and F1 score, providing a comprehensive evaluation of diagnostic capability. We mainly focus on AUROC and F1 score in our main results, and the detailed metric results are in the supplementary material. 

\begin{table*}[t]
\centering
\small
\caption{Demographic characteristics of the full dataset and fine-tuning dataset, summarized per patient and per OCT scan.}
\label{tab:demographics}
\begin{tabular}{@{}llcccc@{}}
\toprule
& & \multicolumn{2}{c}{Full dataset} & \multicolumn{2}{c}{Fine-tuning dataset} \\
\cmidrule(lr){3-4} \cmidrule(lr){5-6}
\textbf{Variable} & \textbf{Category} & \textbf{Per patient, n (\%)} & \textbf{Per OCT, n (\%)} & \textbf{Per patient, n (\%)} & \textbf{Per OCT, n (\%)} \\
\midrule
Gender
& Female   & 18,611 (45.44\%) & 112,950 (48.64\%) & 3,704 (45.41\%) & 23,072 (47.86\%) \\
& Male     & 12,423 (30.33\%) & 80,360 (34.61\%)  & 2,526 (30.97\%) & 17,442 (36.18\%) \\
& Unknown  & 9,922 (24.23\%)  & 38,905 (16.75\%)  & 1,926 (23.61\%) & 7,696 (15.96\%) \\
& Overview & 40,956           & 232,215           & 8,156           & 48,210 \\
\addlinespace[3pt]

Race/Ethnicity
& NHW      & 18,018 (43.99\%) & 117,480 (50.59\%) & 3,641 (44.64\%) & 25,021 (51.90\%) \\
& NHB      & 8,456 (20.65\%)  & 49,669 (21.39\%)  & 1,691 (20.73\%) & 10,288 (21.34\%) \\
& Hispanic & 2,216 (5.41\%)   & 12,746 (5.49\%)   & 423 (5.19\%)    & 2,288 (4.75\%) \\
& Other    & 1,885 (4.60\%)   & 10,810 (4.66\%)   & 380 (4.66\%)    & 2,375 (4.93\%) \\
& Unknown  & 459 (1.12\%)     & 2,605 (1.12\%)    & 95 (1.16\%)     & 542 (1.12\%) \\
& Missing  & 9,922 (24.23\%)  & 38,905 (16.75\%)  & 1,926 (23.61\%) & 7,696 (15.96\%) \\
& Overview & 40,956           & 232,215           & 8,156           & 48,210 \\
\addlinespace[3pt]

Age group
& $<45$        & --- & 23,672 (10.19\%) & --- & 4,326 (8.97\%) \\
& 45--64       & --- & 66,993 (28.85\%) & --- & 14,053 (29.15\%) \\
& 65--74       & --- & 52,368 (22.55\%) & --- & 10,906 (22.62\%) \\
& $\geq 75$    & --- & 50,277 (21.65\%) & --- & 11,229 (23.29\%) \\
& Unknown      & --- & 38,905 (16.75\%) & --- & 7,696 (15.96\%) \\
& Overview     & --- & 232,215          & --- & 48,210 \\
\bottomrule
\end{tabular}
\end{table*}

\noindent Table \ref{tab:demographics} summarizes the demographic characteristics of the full cohort and the fine-tuning dataset at both the patient level and the OCT level. In the full dataset, there are 40,956 patients with 232,215 OCT scans, while the fine-tuning dataset included 8,156 patients and 48,210 OCT scans. The demographic distribution was similar between the two datasets.
For patient sex, females represented the largest subgroup (45\%), followed by males (30\%) and patients without available EHR data (24\%). For race and ethnicity, non-Hispanic White patients accounted for the largest proportion (44\%), followed by non-Hispanic Black patients (20\%), whereas Hispanic, other, and unknown racial or ethnic groups comprised smaller proportions. A substantial proportion of records also lacked EHR data (24\%).
Age-group information was summarized only at the OCT-scan level because individual patients could contribute multiple OCT scans over time. The largest proportion of scans came from patients aged 45–64 years (29\%), followed by those aged 65–74 years (22\%) and those aged 75 years or older (21\%). Overall, the fine-tuning dataset showed a demographic distribution that was largely consistent with that of the full cohort.

\subsection{State-of-the-art diagnostic performance across diverse pathologies}
\begin{figure}[H]
  \centering
  \includegraphics[width=\linewidth]{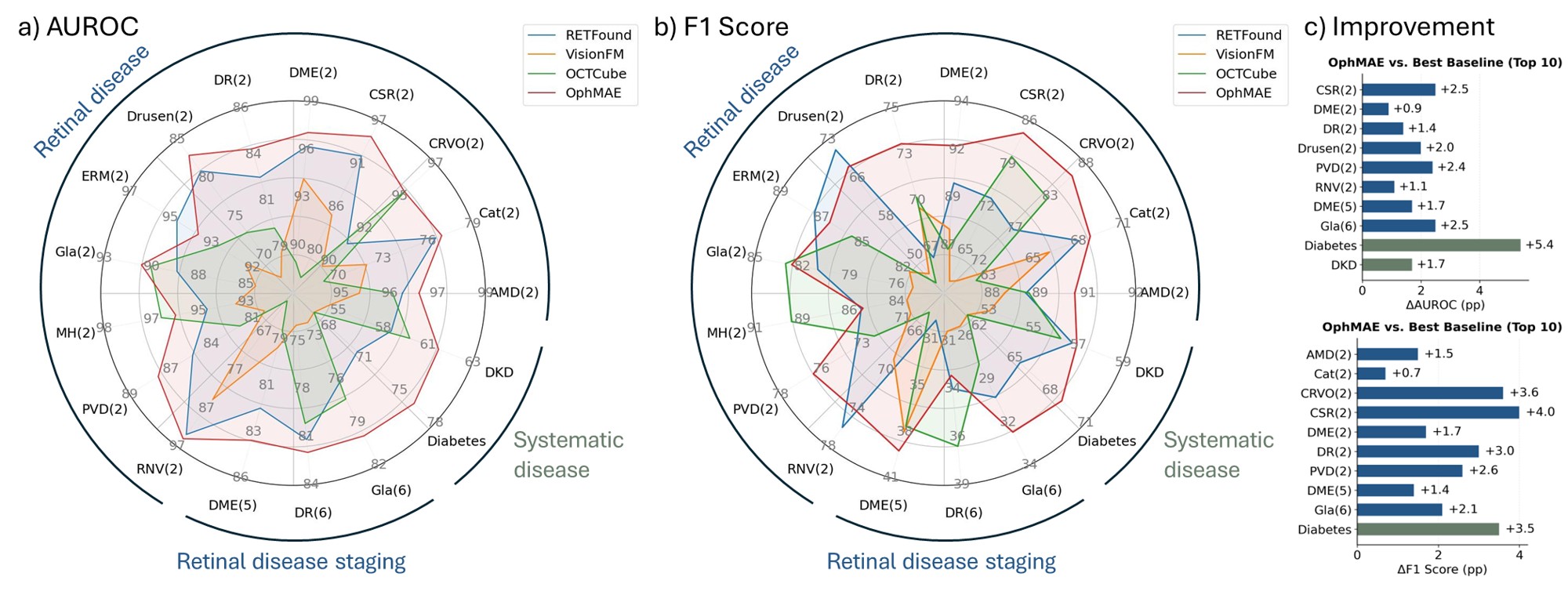}
  \caption{\textbf{Performance comparison of OphMAE and baseline models across evaluated tasks.} The figure summarizes the AUROC (a) and F1 scores (b) of OphMAE and the baseline models on the downstream classification tasks. AUROC was used to assess overall discriminative performance, whereas F1 score was used to evaluate the balance between precision and recall. The results show the relative predictive performance of the proposed multimodal foundation model compared with conventional supervised models and pretrained foundation-model baselines. Subfigure (c) shows the top 10 absolute performance improvements of OphMAE over the strongest competing model for each task, measured in percentage points (pp) for AUROC and F1 score. Abbreviations: Gla, glaucoma; Cat, Cataract. The (2) indicates binary classification (disease detection), whereas (5) and (6) indicate multi-class classification (disease staging).}
  \label{fig:aurocf1}
\end{figure}

\noindent  In a rigorous comparative analysis against leading foundation models (Figure \ref{fig:aurocf1}), including the single-modality RETFound and the multimodal pretrained VisionFM, the proposed OphMAE demonstrated superior diagnostic accuracy across the vast majority of evaluated tasks. In the detection of AMD, a condition where the complementary views of En face OCT and OCT are critical for identifying drusen and atrophy, OphMAE achieved the best Area Under the Curve (AUC) of 96.9\% and an F1-score of 90.5\%. This performance significantly surpasses that of RETFound (AUC 96.4\%) and VisionFM (AUC 95.1\%), confirming the value of our specialized pre-training objectives. Similarly, for the binary classification of Diabetic Macular Edema (DME), our model reached an AUC of 97.2\%, markedly outperforming VisionFM (94.2\%) and the task-specific baseline OCTCube (88.8\%). 

\noindent The advantages of our architecture were most pronounced in detecting pathologies with complex vascular or surface manifestations. For RNV, OphMAE achieved an AUC of 94.7\%, whereas the structural OCT–only SOTA OCTCube achieved only 59.4\%. This large performance gap highlights the limitation of relying solely on cross-sectional B-scans for vascular pathology detection and underscores the importance of incorporating en face OCT information. OphMAE also demonstrated strong generalization in Glaucoma (AUC 90.9\%) and CSR (AUC 95.0\%), consistently outperforming competing models. Among the comparator foundation models, RETFound generally achieved the strongest performance across most tasks, whereas OCTCube and VisionFM performed less well on our cohort. Overall, these support the effectiveness of our 3D-2D multimodal design for ophthalmic foundation model pre-training. 

\noindent For systemic disease, OphMAE showed the strongest AUROC and F1 score for both diabetes and DKD.In contrast to retinal disease diagnosis, where most tasks achieved AUROCs of 80-90\% AUROC, performance for systemic disease prediction was lower overall, with binary classification AUROCs generally in the 60-70\% range. This result is expected, as systemic diseases are not directly defined by retinal abnormalities, and retinal imaging likely reflects only part of the disease process, while other systemic factors may also influence ocular appearance. Still, the diabetes prediction performance (AUROC, 76.1\%) suggests that diabetes is associated with measurable retinal changes, consistent with the well-established relationship between diabetes and retinal diseases such as DR and DME.

\noindent The Figure \ref{fig:aurocf1} further quantifies where OphMAE provided the greatest benefit over the best competing model. For AUROC, the largest improvement was observed for diabetes prediction, with a 5.4\% gain, followed by CSR, Glaucoma staging, proliferative DR, Drusen, DKD, and DME. For F1 score, the largest gains were observed for CSR, CRVO/CRAO, diabetes, DR, PVD, Glaucoma staging, and DME. These improvements indicate that OphMAE was not only competitive in tasks where existing models already performed strongly, but also provided meaningful gains in tasks requiring integration of retinal imaging signals.

\subsection{Ablation study of cross-modal fusion designs} 

\noindent To assess the specific benefit of integrating en face OCT with OCT, we evaluated state-of-the-art models under both single-modality and dual-modality inference settings. In the single-modality setting, each model was fine-tuned using either OCT or en face OCT alone. In the dual-modality setting, the final prediction was obtained by combining the outputs of the two separately fine-tuned single-modality models. For RETFound and VisionFM, we used the fundus-pretrained variants, RETFound\_mae\_natureCFP and VFM\_Fundus, respectively, whereas for OCTCube, we used OCTCube-IR. Dual-modality inference was implemented using a simple late-fusion strategy that averaged the classification outputs from the OCT and en face OCT models on the test set.

\begin{figure}[H]
  \centering
  \includegraphics[width=0.85\linewidth]{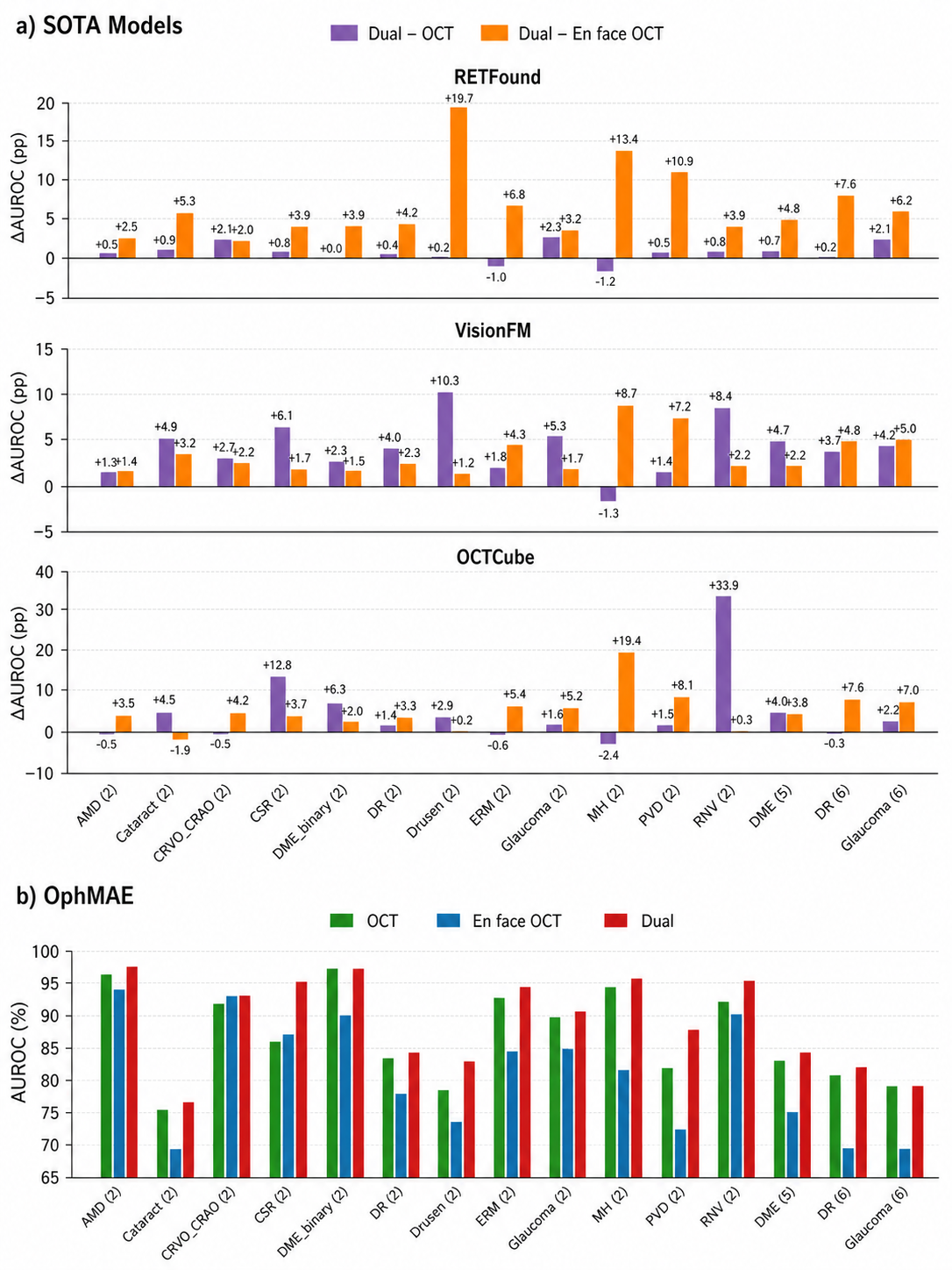}
  \caption{\textbf{AUROC comparison of proposed state-of-the-art and OphMAE models under different input modality settings.} a) The figure demonstrates the AUROC improvement of multi-modality in state-of-the-art foundation models across the downstream tasks. b) The figure summarizes the AUROC achieved by proposed OphMAE model across the downstream tasks when using different imaging modalities as input. We calculate the performance improvement as dual modality performance minus single modality performance. Abbreviations: SOTA: state-of-the-art.  The (2) indicates binary classification (disease detection), whereas (5) and (6) indicate multi-class classification (disease staging).}
  \label{fig:ab_multimodal_combined}
\end{figure}

\noindent Figure \ref{fig:ab_multimodal_combined} (a) illustrates the performance improvement while using dual-modality input for the state-of-the-art foundation models. Overall, dual-input inference achieved positive improvement compared to single modality, especially compared with en face–only. The results confirm that the "Dual-Input" configuration consistently yields the highest performance, particularly for pathologies with distinct surface features that can be diagnosed with both modalities. (e.g., Glaucoma, CSR, DME) For instance, the "Dual Input" achieves about 4\% AUROC improvement in Glaucoma, CSR, and DME tasks.
By contrast, dual-input setting results in negative improvement compared to OCT-only in MH tasks, with about 1\% decrease of AUROC. This finding may be explained by the fact that MH presents clear structural abnormalities on OCT, whereas these features may be less visible on grayscale en face images. In sum, these findings support the stability and strength of multi-modality relative to single-modality models.

\subsection{Adaptive inference with robustness in resource-limited settings} 

\noindent A pivotal engineering contribution of OphMAE is its adaptive inference capability, which maintains clinical utility even when imaging equipment is limited or malfunctioning. When the OCT input stream was withheld to simulate a resource-constrained clinic equipped only with 2D cameras, OphMAE adapted seamlessly to the missing modality. The results are shown in Figure \ref{fig:ab_multimodal_combined} (b).Under this en face image-only setting, the model maintained a highly respectable diagnostic accuracy, achieving an AUC of 93.7\% for AMD detection. This represents a slight degradation, standing in sharp contrast to rigid multimodal architectures that fail catastrophically when an input channel is missing. This capability allows a single, unified model to be deployed across a stratified healthcare system—delivering maximum precision in tertiary centers utilizing full multimodal capabilities, while retaining robust screening-level accuracy in primary care centers relying solely on 2D imaging.


\subsection{Data efficiency and fairness considerations}

\noindent Foundation models are expected to be more data efficient because pre-training can provide rich and transferable representations. We evaluated this hypothesis by fine-tuning OphMAE on reduced training subsets of 500, 300, and 100 samples drawn from the original training and validation data. To reduce selection bias, we repeated the sampling process with 10 different random seeds. The held-out test set was kept identical to that used in the full-dataset experiments. Model performance was evaluated using test AUROC with 95\% confidence intervals.

\begin{figure}[H]
  \centering
  \includegraphics[width=\linewidth]{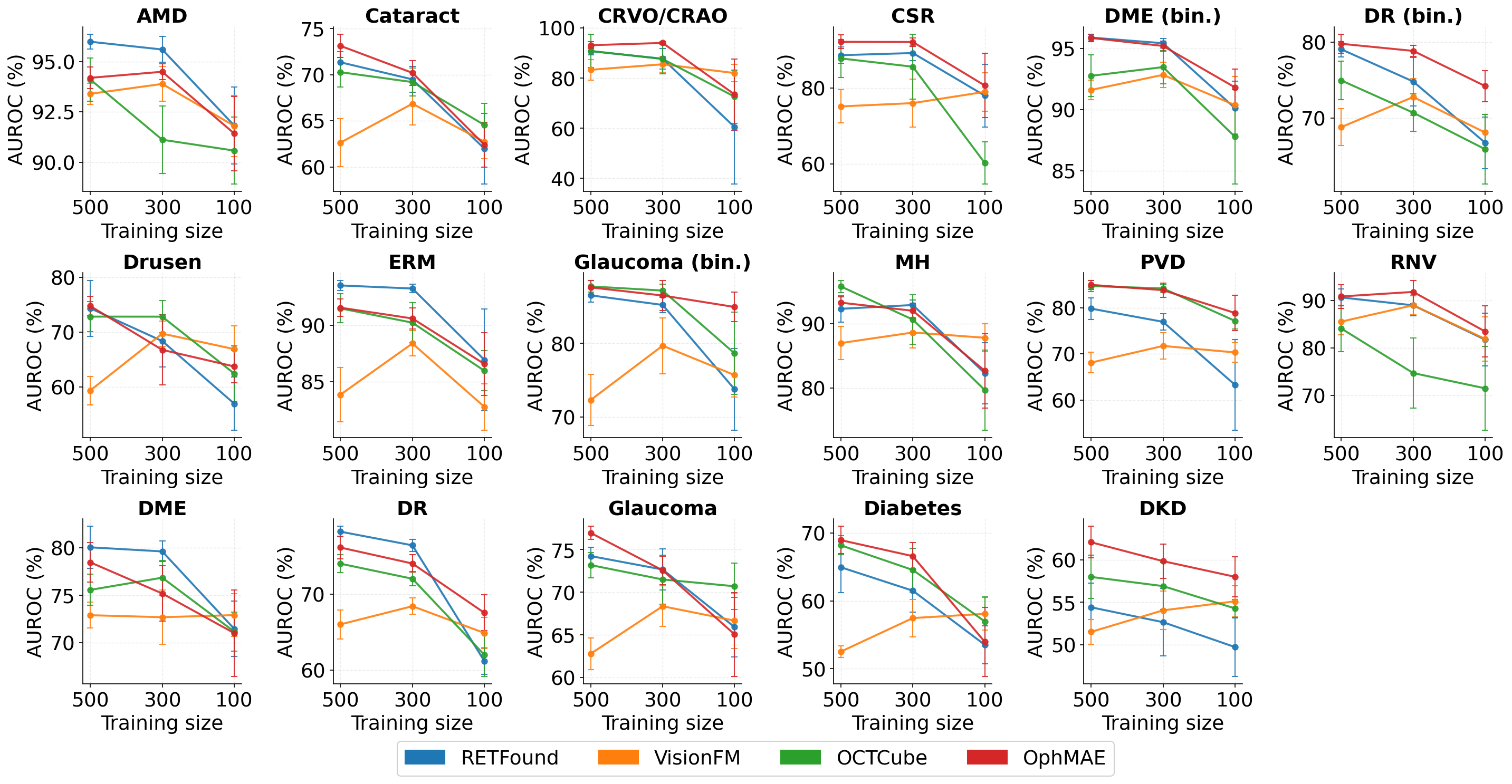}
  \caption{\textbf{AUROC of OphMAE and baseline models across different training set sizes.} The figure summarizes the downstream classification performance of OphMAE and the baseline models under subset fine-tuning settings. Test AUROC is shown for each model when the number of training samples per class was progressively reduced, enabling comparison of performance robustness under limited-data conditions. Error bars indicate variability across repeated experiments and represent 95\% confidence intervals estimated from 10 random sampling seeds.}

  \label{fig:subset_boot}
\end{figure}

\noindent As shown in Figure \ref{fig:subset_boot}, OphMAE preserved performance well as training data decreased and showed particular strength in several low-data settings. Compared with RETFound, OphMAE achieved higher AUROC across training size in most tasks. For instance, OphMAE achieves the best result in CSR, DR binary, and RNV tasks with around 3\% improvement compared to the second-best model. Also, in the extreme low-data regime of 100 training samples for Glaucoma detection, OphMAE maintained an AUC of 83\%, whereas the supervised baseline OCTCube failed to learn effectively, yielding an AUC of 71\%. Overall, our proposed OphMAE demonstrates more stable performance when training datasets are limited.

\noindent Among the SOTA foundation models, RETFound achieves superior performance across different tasks. For instance, RETFound performed better for ERM and multiclass DME in several settings. The 2D image has much lower dimension, making the model easier to fine-tune in the size-limited dataset. OCTCube has a similar trend of performance changes with OphMAE and RETFound while training size becomes lower in most tasks, but the overall performance is lower than these models.

\noindent By comparing the full dataset and the subset training performance, we found that test AUROC is declining while the training data is decreasing in most tasks and models, which matches our intuitive concept of deep learning. However, the declining rate is different, from the full dataset to 300 training samples, the performance slightly declined and even stayed similar, but when reduced to 100 training samples, the model performance was heavily damaged in most tasks. This experiment illustrates the stability of foundation models with limited finetuned data.

\noindent To assess potential bias in the foundation model, we performed a subgroup fairness analysis across key demographic variables, including age, sex, and race/ethnicity. For each variable, patients were divided into a protected group and a privileged group. We then evaluated subgroup performance using multiple fairness-related metrics, including positive predictive value (PPV), false positive rate (FPR), true positive rate (TPR), negative predictive value (NPV), the false negative-to-false positive ratio (FN/FP) as a measure of treatment equality, false negative rate (FNR), overall accuracy (ACC), and area under the receiver operating characteristic curve (AUROC). To summarize subgroup disparity, we calculated the ratio between the protected and privileged groups, where a ratio of 1.0 indicates parity between groups. By comparing these metrics across subgroups, we assessed whether the model showed systematic differences in prediction reliability, missed detection rate, and overall discrimination performance. Larger deviations from 1.0 were considered indicative of greater potential disparity.

\noindent For age, we defined patients aged 75 years or older (22\%) as the protected group and those aged 45–64 years (29\%) as the privileged group. For sex, females (45\%) were defined as the protected group and males (30\%) as the privileged group. For race/ethnicity, non-Hispanic Black patients (20\%) were defined as the protected group and non-Hispanic White patients (44\%) as the privileged group. Because the intended use of our foundation model is early eye disease screening, we focused primarily on FPR as the main fairness metric, as this measure reflects the rate of false alarms across subgroups. Results for the remaining metrics and their descriptions are provided in the Supplementary Materials.

\noindent When examining FPR across sensitive subgroups, age showed the largest subgroup disparity. In particular, the protected group aged 75 years or older often had substantially higher FPR than the privileged group aged 45–64 years, with the ratio exceeding 1.5 in several tasks. The largest age-related disparities were observed for PVD, AMD, and glaucoma, where the FPR differences exceeded fourfold. This pattern is likely related to the strong age dependence of these diseases. 
For sex, the ratios are generally less than 1.0, which indicates that protected groups (females) achieve more prediction reliability than males. This pattern may partly reflect differences in subgroup representation in the fine-tuning cohort, where females constituted a larger proportion of patients than males (45\% female v.s. 31\% male). Drusen was the main exception, for which females showed a higher FPR. Compared with RETFound, OphMAE generally achieved FPR ratios closer to 1.0 across most tasks, suggesting a better balance between demographic groups. This pattern was particularly evident for AMD (OphMAE: 1.1; RETFound: 0.6), ERM (OphMAE: 0.8; RETFound: 0.5), and Drusen (OphMAE: 1.2; RETFound: 1.6). 
In the race/ethnicity analysis, most ratios were below 1.0, and OphMAE showed better or competitive FPR parity in most tasks, indicating more balanced prediction reliability across groups. Among all tasks, AMD showed the lowest race-based FPR ratio for both OphMAE and RETFound (0.35 and 0.32, respectively), and MH and Glaucoma have the highest FPR fairness ratio (OphMAE: 1.04 and 1.72; RETFound: 1.34 and 1.84). These findings suggest substantial task-specific heterogeneity in subgroup fairness. Overall, the subgroup analysis revealed measurable disparities between protected and privileged groups, while OphMAE generally showed more balanced subgroup performance than RETFound.

\begin{figure}[H]
  \centering
  \includegraphics[width=\linewidth]{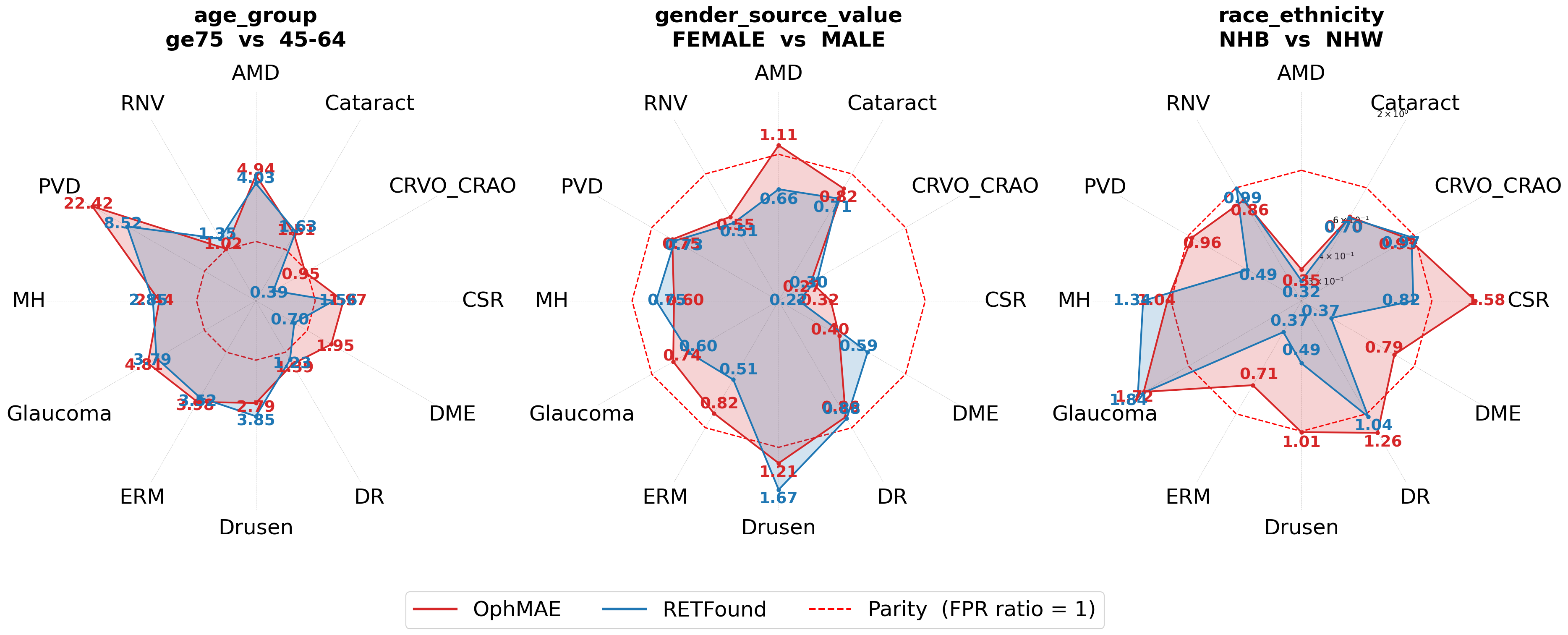}
  \caption{\textbf{FPR-based subgroup fairness comparison between OphMAE and RETFound.} The figure summarizes the false-positive rate (FPR) ratios between protected and privileged groups across age, sex, and race/ethnicity subgroups for OphMAE and RETFound. A ratio closer to 1.0 indicates better subgroup parity, whereas larger deviations from 1.0 indicate greater disparity in false-positive performance. }
  \label{fig:fairness}
\end{figure}

\section{Generalization to public fundus and 3D OCT dataset}

\begin{figure}[H]
  \centering
  \includegraphics[width=0.8\linewidth]{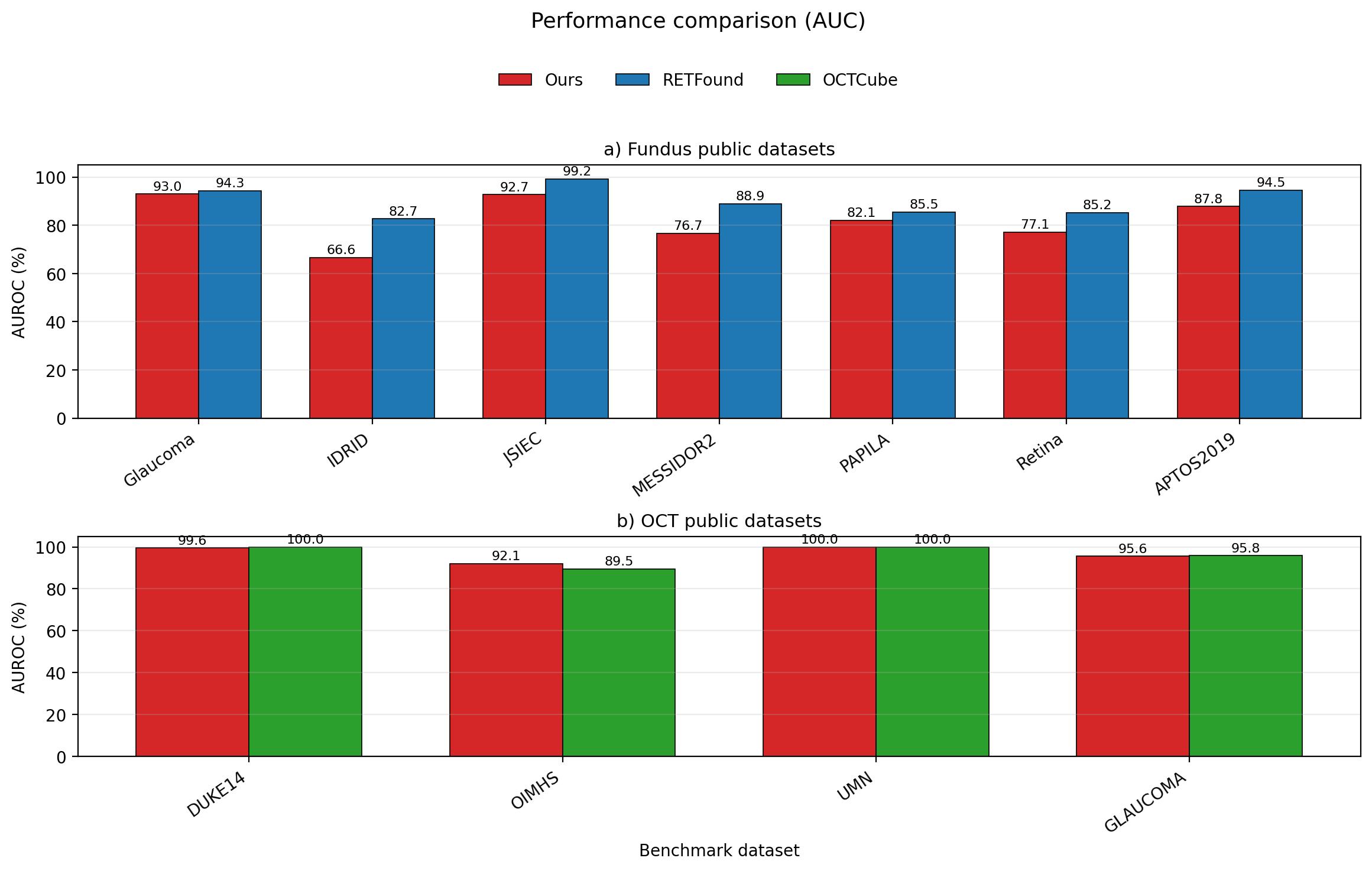}
  \caption{\textbf{AUROC performance of OphMAE and comparator foundation models on public single-modality benchmarks.} \textbf{a,} Public fundus datasets. \textbf{b,} Public 3D OCT datasets. For each benchmark, the corresponding modality-specific model was fine-tuned and evaluated. AUROC is reported to assess discriminative performance across datasets.}

  \label{fig:pub_aurocf1}
\end{figure}

To further investigate the effectiveness of the proposed OphMAE modality-specific pretrained feature extractors, we evaluated OphMAE and SOTA models (RETFound, OCTCube) on the public 2D fundus and 3D OCT datasets. As these datasets are single-modality datasets, containing either 2D fundus images or 3D OCT volumes, only the corresponding feature extractor was used for fine-tuning and evaluation. To ensure fair comparison, we followed the RETFound experimental settings, including data splits, for the public fundus datasets and the OCTCube experimental settings for the public 3D OCT datasets.

\noindent As shown in Figure \ref{fig:pub_aurocf1}, we evaluated our proposed model across multiple fundus and OCT benchmarks using AUROC (\%). On fundus datasets, RETFound CFP consistently outperformed our finetuned OphMAE, with an AUROC gap of approximately 1.3--16.1\% across benchmarks. This degradation is expected due to modality mismatch: OphMAE is pretrained on grayscale en face OCT, whereas these public benchmarks are color fundus photographs. In contrast, RETFound CFP is pretrained directly on fundus images, providing stronger domain alignment.
On the OCT benchmarks, OphMAE achieved consistently high AUROC and remained competitive with OCTCube. In particular, OphMAE outperformed OCTCube on OIMHS (92.1 vs. 89.5 AUROC) and matched performance on UMN (100.0 AUROC for both). On DUKE14 and the OCT glaucoma benchmark, OphMAE was slightly lower ($<$ 0.5\% absolute AUROC) than OCTCube, indicating comparable transfer with minor performance loss.
AUROCs on the public OCT benchmarks were generally higher than those observed in our UF real-world cohort, underscoring the greater difficulty of real-world EHR-linked data. This difference likely reflects the combined effects of heterogeneous imaging protocols, more diverse patient populations, and increased label noise.

\section{Discussion}


This study introduces a novel multimodal foundation model that integrates 3D OCT and 2D en face imaging for enhanced diagnosis of retinal and systemic diseases. The proposed model addresses fundamental limitations of existing single-modality approaches by leveraging complementary information from multiple imaging modalities. While OCT excels at capturing microstructural changes in the retina, en face OCT provides critical information about choroidal vasculature and pigmentary changes, making the combination of these modalities essential for comprehensive diagnostic evaluation.
Integrating complementary imaging modalities within a foundation model is important for accurate ophthalmic disease screening. Our low-data and resource-constrained experiments show that OphMAE retains strong downstream diagnostic performance even when fine-tuned with only a few hundred samples or when deployed under incomplete multimodal input settings. Together, these results support the robustness and practical clinical utility of OphMAE in real-world environments, where labeled data are often limited and multimodal imaging may be incomplete.


\noindent In clinical practice, ophthalmologists utilize multiple imaging modalities to diagnose eye diseases. A common combination is optical coherence tomography (OCT) and en face imaging \cite{enface_app1,enface_app2,enface_app3}, which provides complementary information for detailed assessment of retinal lesions. OphMAE was designed to reflect this multimodal clinical workflow by jointly learning from OCT and en face imaging. To better align the two modalities during pretraining, we further introduced cross-modal relation masking and a multi-space semantic similarity objective. In experiments across diverse pathologies, OphMAE consistently outperformed state-of-the-art OCT foundation models, including RETFound, VisionFM, and OCTCube, on most tasks. Moreover, the cross-modal fusion ablation study showed that multimodal integration improved performance even when using a simple late-fusion strategy at the prediction level. In sum, these findings support the clinical rationale for incorporating complementary imaging modalities into foundation models and highlight the effectiveness of the proposed OphMAE multimodal framework.



\noindent Our data efficiency experiments further highlight the value of multimodal imaging for ophthalmic diagnosis. The largest gains from multimodal input were observed in tasks such as Glaucoma, CSR, and DME, which demonstrate the multimodal diagnosis criteria in ophthalmology guidelines or reports. \cite{glaucoma_guide,csr_guide,dr_guide} On the other hand, the OCT-only model outperforms the multi-modality model in MH or ERM tasks, where their related clinical reports clearly note that OCT is the gold standard of diagnosis. \cite{mh_oct,erm_guide} Furthermore, the superiority of the OCT modality in diagnosing these specific conditions stems from the fact that their pathological features are predominantly depth-resolved, rendering them obscure on planar grayscale en face imaging but highly conspicuous on cross-sectional OCT (e.g., lamellar macular defects in MH). \cite{mh_sign} 
Comparing the single modality result, we can find out that the OCT-only model reaches higher performance in RETFound and OCTCube, but often achieves lower performance in VisionFM. The superior performance in the OCT modality also matches the research trend that utilizes OCT as a major tool for retinal-related diagnosis. Overall, the experiment results reveal the multi-modality advantages across SOTA foundation models, which further support the multi-modality claims in OCTCube.

\noindent The data efficiency analysis showed that the performance of all foundation models declined only modestly when the training set was reduced to 300 samples, but dropped more markedly in the 100-sample setting. Among the evaluated methods, OphMAE generally maintained the strongest performance as training data decreased, while RETFound remained competitive in several tasks. Although 2D modalities may retain some advantage under limited-data conditions because of their lower input dimensionality, OphMAE was able to preserve strong performance despite using multimodal input, suggesting that pre-training can mitigate the additional complexity of multimodal learning. 
The wider confidence intervals observed in the 100-sample setting further underscore the instability of training under severe data scarcity. Even so, OphMAE remained among the best-performing methods across tasks, supporting its robustness in data-limited settings. Taken together, these findings suggest that OphMAE is effective under standard training conditions and may serve as a useful reference for downstream fine-tuning in other ophthalmic diseases where labeled multimodal datasets are limited.



\noindent Our subgroup analysis identified measurable differences in false positive rate (FPR) across demographic groups, with age showing the largest disparities. In particular, patients aged 75 years or older often had higher FPR than those aged 45–64 years. The largest age-related disparities were observed for PVD, AMD, and glaucoma. This pattern is clinically plausible, as these conditions are strongly age-associated, \cite{amd_guide,glaucoma_guide}, and older patients are also more likely to have concurrent age-related ocular changes that may complicate image interpretation and increase the likelihood of false positive predictions. AMD is a prototypical age-related retinal disease, glaucoma prevalence also increases with age, and PVD becomes markedly more common in older adults. \cite{impact_age,pvd_age}
Sex-based differences were generally smaller. In most tasks, the female-to-male FPR ratio was below 1.0, suggesting lower false positive rates in females than in males, with drusen as an exception. One possible explanation is an imbalance in subgroup representation during fine-tuning, as females comprised a larger proportion of the cohort than males. At the same time, sex effects in ophthalmic disease are known to be heterogeneous across conditions. \cite{2024gender} 
For race and ethnicity, AMD showed the lowest race-based FPR ratio, whereas MH and glaucoma showed the largest departures from parity. These results suggest that fairness in ophthalmic foundation models may vary substantially by disease context rather than following a single consistent pattern across all tasks.

\noindent Overall, subgroup fairness varied by demographic factor and disease task, with age showing the largest disparities in false positive rates. Although measurable differences remained across age, sex, and race/ethnicity groups, OphMAE generally demonstrated more balanced FPR performance than RETFound in several tasks. These findings highlight the importance of evaluating ophthalmic foundation models with subgroup-specific error metrics, particularly for screening applications where unequal false positive burden may affect clinical utility.


\noindent The clinical relevance of this work lies in its potential to support earlier screening of a variety of ophthalmic diseases. By achieving strong performance in both single-modality and multimodal settings, the model shows promise for use across a range of clinical environments, including settings where imaging availability may vary. Its favorable external validation performance further suggests that the learned representations are not limited to a single institution, supporting the potential utility of this approach in broader real-world deployment.

\noindent This study has several limitations. First, the training data were derived predominantly from a single imaging platform (Heidelberg Spectralis), which may limit generalizability to images acquired from other devices or acquisition settings. Broader validation across imaging vendors and clinical workflows will therefore be important. Second, although we evaluated multiple retinal diseases, the current diagnostic scope remains incomplete. Extending this framework to additional ophthalmic conditions, including optic nerve and inflammatory disorders, would help clarify its broader clinical utility. Third, computational efficiency was not a primary focus of the present study, and further optimization of inference speed and deployment cost will be necessary for routine clinical implementation.

\noindent Several directions warrant future investigation. Cross-platform and multi-center validation will be essential to assess robustness under more heterogeneous real-world conditions. Future work could also explore broader task formulations, such as multi-task learning for diagnosis, progression risk, and treatment response prediction. In addition, more systematic interpretability analyses may help determine how multimodal foundation models can provide clinically meaningful evidence alongside predictions. Or more fairness analyses to make sure the model was not biased against some sensitive subgroups. Overall, our findings support multimodal foundation modeling as a promising direction for ophthalmic AI, while underscoring the importance of broader validation, efficiency optimization, and clinically grounded interpretation before real-world deployment.






%
\section{Methods}

\begin{figure}[H]
  \centering
  \includegraphics[width=0.85\linewidth]{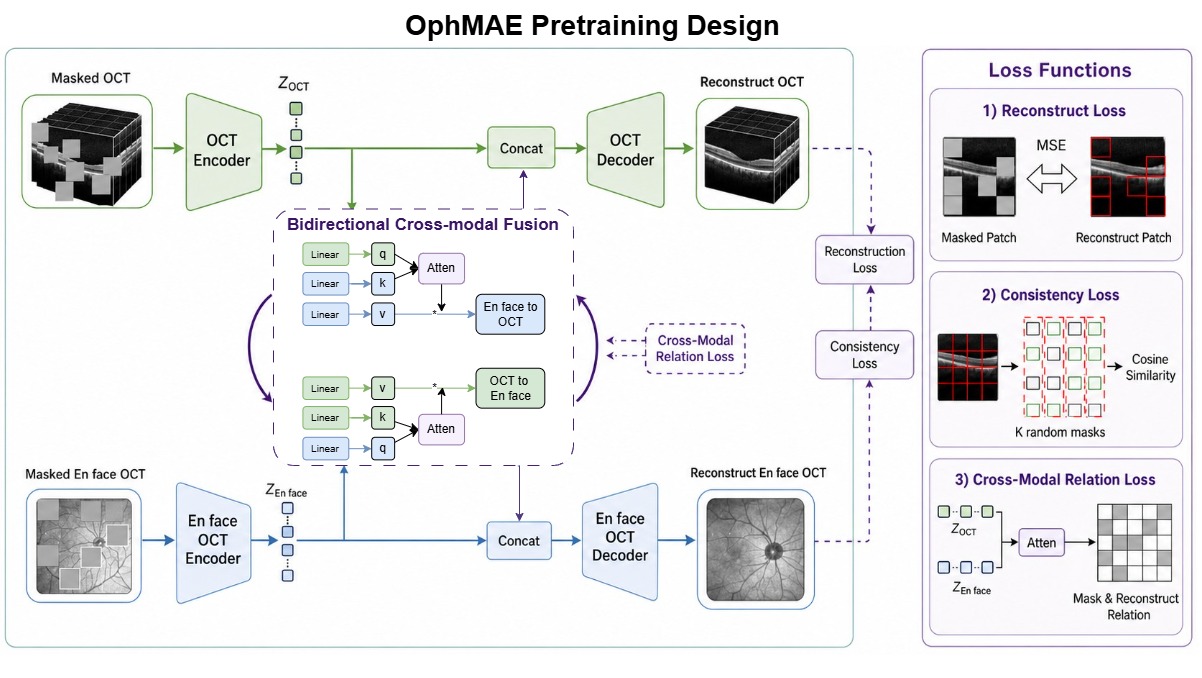}
  \caption{\textbf{Architecture and self-supervised pre-training strategy of OphMAE.} The proposed OphMAE framework consists of modality-specific encoders for 3D volumetric OCT and 2D en face OCT, a bidirectional cross-modal fusion module, and lightweight modality-specific decoders used only during pre-training. The model is optimized with three objectives: modality-specific reconstruction loss, multi-space semantic similarity consistency across multiple masking views, and cross-modal relation loss in the feature space.}
  \label{fig:model_arch}
\end{figure}

\subsection{Multi-imaging foundation model architecture}
\noindent Our OphMAE framework builds upon the masked autoencoder (MAE) paradigm, extending it to accommodate both 3D volumetric OCT data and 2D en face OCT through a unified pre-training framework. The architecture consists of three primary components: modality-specific encoders for both 3D volumetric OCT data and 2D en face OCT, a bidirectional cross-modal fusion module, and modality-specific decoders strictly reserved for the reconstruction tasks during pre-training.

\noindent Formally, the architecture is designed to process a 3D OCT volume $X_{\rm OCT} \in \mathbb{R}^{H \times W \times D}$ and a paired 2D en face OCT image $X_{\rm IR} \in \mathbb{R}^{H \times W}$. For volumetric OCT processing, we employed a 3D Vision Transformer (ViT) encoder that partitions  $X_{\rm OCT}$ into non-overlapping 3D patches (16 × 16 × 5 voxels). These patches are linearly projected into a high-dimensional embedding space $Z_{\rm OCT} \in \mathbb{R}^{N_{1} \times C}$, augmented with learnable 3D positional encodings to preserve spatial relationships within the volumetric data. Simultaneously, the planar branch utilizes a standard 2D ViT encoder to process the 2D en face OCT image $X_{\rm IR}$ into $16 \times 16$ pixel patches with corresponding positional embeddings, mapped to $Z_{\rm IR} \in \mathbb{R}^{N_{2} \times C}$.

\noindent The bidirectional cross-modal fusion module represents the core architectural design, explicitly engineered to enable bidirectional information exchange between the intermediate layers of the two modality-specific encoders. Rather than concatenating features at the final layer, this intermediate fusion mechanism employs a bi-directional cross-attention operation. Specifically, the latent representations from the OCT encoder serve as queries ($Q$) to attend to the keys ($K$) and values ($V$) generated from the en face OCT encoder, and vice versa. This formulation allows the network to dynamically resolve anatomical ambiguities in one modality by fetching spatially correlated structural context from the other, utilizing modality-specific normalization layers and learnable embedding vectors to bridge the inherent dimensionality gap between volumetric and planar data.

\noindent  In alignment with the standard MAE paradigm, the decoding phase utilizes an asymmetric architecture featuring two independent, lightweight ViT decoders: a volumetric decoder for OCT and a planar decoder for IR. Following the encoding and bidirectional cross-modal fusion stages, the input to each modality-specific decoder is constructed by combining the encoded visible tokens with a set of shared, learnable [MASK] tokens representing the dropped patches. To restore spatial awareness prior to reconstruction, modality-specific positional embeddings are systematically added to this full set of tokens. The volumetric decoder then maps the latent representations back to the original $16 \times 16 \times 5$ voxel resolution, while the planar decoder reconstructs the missing $16 \times 16$ pixel patches. By intentionally designing these decoders to be significantly shallower and narrower than their corresponding encoders, the architecture forces the encoders to learn highly abstracted, semantically rich representations, while restricting the decoders to the localized task of pixel and voxel reconstruction during the pre-training phase.

\subsection{Self-supervised pre-training strategy}
\noindent Our pre-training approach optimizes a composite objective function that transcends conventional masked modeling by harnessing the rich, complementary supervisory signals inherent in paired OCT-IR datasets. The overall optimization objective is defined as $\mathcal{L}_{total} = \lambda_{1}\mathcal{L}_{recon} + \lambda_{2}\mathcal{L}_{cross\_relation} + \lambda_{3}\mathcal{L}_{consistency}$, where the hyperparameters $\lambda_{1}, \lambda_{2}, \lambda_{3}$ balance the relative contributions of localized pixel reconstruction, bidirectional cross-modal topological alignment, and global semantic consistency. To establish robust intra-modal representations, we implement the Modality-Specific Reconstruction with an independent reconstruction loss ($\mathcal{L}_{recon}$) where a high proportion (75\%) of patches within each modality are randomly masked and subsequently reconstructed by modality-specific decoders. This standard Mean Squared Error (MSE) objective forces the encoders to capture fundamental structural topologies, such as retinal layer boundaries in OCT and vascular networks in IR, maintaining the proven representational efficacy of the MAE paradigm.

\subsection{Cross-modal relation masking at the semantic level}
\noindent A critical technical contribution of OphMAE is the introduction of Cross-Modal Relation Masking ($\mathcal{L}_{cross\_relation}$), designed specifically to leverage multi-imaging knowledge at the semantic level. Unlike standard masked autoencoders that operate exclusively on pixel-space reconstruction, this module computes the relational dynamics between the two modality branches directly within the feature space. Operationally, this methodology entails two distinct forward passes. In the first forward pass, we conduct a complete computation using all unmasked tokens to calculate the relation between the two modalities through the cross-attention mechanism within the bidirectional cross-modal fusion module, yielding a dense, uncorrupted relational matrix. In the second forward pass, we deliberately erase a subset of these computed relations. The network is then tasked with reconstructing the complete relation matrix from this masked state within the decoder space. The training objective for this semantic reconstruction is formulated using a Mean Square Error (MSE) loss, minimizing the discrepancy between the predicted relations and the complete relations obtained from the first pass. By enforcing this relation reconstruction in the feature space rather than solely relying on pixel-space puzzle-solving, we supplement the model with significantly more supervision information, ensuring a far more sufficient and robust training objective.

\subsection{Multi-space semantic similarity consistency}
\noindent To directly address the notoriously inefficient learning dynamics and limited supervision signals of standard MAE architectures, which typically generate only a single set of random masks per sample per iteration, we introduce a Multi-Space Semantic Similarity Consistency objective ($\mathcal{L}_{consistency}$). This module is engineered to extract maximally abundant pre-training supervision from each individual sample. Rather than relying on a single masking view, we apply $K$ independent sets of random masks to the same input sample simultaneously during the pre-training execution. This strategy intrinsically generates $K$ times the volume of supervision information per forward iteration, vastly improving data utilization efficiency. Furthermore, these $K$ independent masking views naturally produce overlapping masked regions across the different mask sets. We enforce a strict semantic consistency constraint across these overlapping areas, ensuring that the reconstructed predictions for the same spatial region remain invariant regardless of the specific mask configuration. By minimizing the divergence among the $K$ predictions in these overlapped regions, this multi-space consistency approach provides dense, high-quality supervisory signals, drastically addressing the inefficient learning bottleneck of traditional MAE models.

\subsection{Adaptive inference mechanisms}
\noindent To accommodate the variable availability of imaging hardware across stratified clinical environments, our model is engineered with an adaptive inference capability that maintains high performance regardless of input completeness. The architecture seamlessly reconfigures its data routing paths during deployment. For multimodal scenarios where both scans are available, the full architecture, including the bidirectional cross-modal fusion module, is activated to synthesize complementary features. Crucially, when confronted with single-modality inputs (e.g., an isolated en face OCT image in a community screening setting), the forward pass automatically bypasses the bidirectional cross-modal fusion block. The signal is routed exclusively through the corresponding single encoder (e.g., the en face OCT Encoder), producing a latent representation that is directly digested by the downstream task head. This dynamic structural plasticity ensures robust predictive stability without requiring modality-specific retraining or the introduction of zero-padding artifacts, effectively mitigating catastrophic failure when an input channel is compromised by equipment malfunction or poor patient compliance.

\subsection{Downstream task fine-tuning and annotation scarcity handling}
\noindent Following the self-supervised pre-training phase, the foundation model was systematically fine-tuned to tackle 17 distinct downstream diagnostic classifications. Recognizing the extreme annotation scarcity prevalent in specialized medical domains, our fine-tuning protocol was explicitly optimized for data efficiency. For few-shot scenarios (e.g., 100 or 500 samples), we initialized the task-specific classification heads while strictly freezing the pre-trained encoder weights and the bidirectional cross-modal fusion module to prevent catastrophic forgetting and overfitting. Subsequently, an end-to-end fine-tuning phase was executed utilizing a lower learning rate bounded by a cosine annealing schedule with a linear warm-up phase. We employed the AdamW optimizer with rigorous weight decay regularization to ensure that the robust semantic topologies established during pre-training were smoothly transferred to the specific decision boundaries required for accurate disease stratification.

\section{Evaluation metrics}
To comprehensively assess the diagnostic performance of our multimodal foundation model across various ophthalmic conditions, we employed a comprehensive set of classification metrics that capture different aspects of model performance and clinical utility. Specifically, model performance was assessed using area under the receiver operating characteristic curve (AUROC), Area Under the Precision-Recall Curve (AUC-PR), accuracy, F1 score, precision, recall, Cohen’s kappa, and Matthews correlation coefficient (MCC).

\noindent Area Under the Receiver Operating Characteristic Curve (AUC-ROC) served as our primary evaluation metric, providing a threshold-independent measure of the model's ability to discriminate between positive and negative cases across all possible classification thresholds. AUC-ROC values range from 0.5 (random performance) to 1.0 (perfect discrimination), with values above 0.8 generally considered indicative of good diagnostic performance in medical applications. Area Under the Precision-Recall Curve (AUC-PR) was employed as a complementary metric, particularly valuable for evaluating performance on imbalanced datasets where the prevalence of positive cases may be low. Unlike AUC-ROC, AUC-PR is more sensitive to performance on the minority class and provides insights into the precision-recall trade-offs across different operating points.

\noindent Accuracy was calculated as the proportion of correctly classified instances among all predictions, providing an intuitive measure of overall model performance. However, given the potential for class imbalance in clinical datasets, accuracy was interpreted alongside other metrics to ensure comprehensive performance assessment. F1-Score was computed as the harmonic mean of precision and recall, as F1 = 2 × precision × recall / (precision + recall), providing a balanced measure that accounts for both false positives and false negatives. This metric is particularly valuable in clinical contexts where both missed diagnoses and false alarms carry significant consequences.

\noindent Recall (Sensitivity) was measured as the proportion of actual positive cases correctly identified by the model (true positives / (true positives + false negatives)). High recall is crucial in screening applications where missing positive cases could lead to delayed treatment and vision loss. Precision (Positive Predictive Value) was calculated as the proportion of predicted positive cases that were truly positive (true positives / (true positives + false positives)). High precision is important for minimizing unnecessary follow-up procedures and patient anxiety from false positive diagnoses. Matthews Correlation Coefficient (MCC) was included as a robust metric that accounts for class imbalance and provides a balanced measure of prediction quality across all four confusion matrix categories. MCC values range from -1 to +1, where +1 indicates perfect prediction, 0 represents random performance, and -1 indicates complete disagreement between predictions and observations. Cohen’s kappa was used to quantify agreement between model predictions and reference labels beyond chance.

\noindent For multi-class classification tasks, macro-averaged versions of precision, recall, and F1-score were computed by calculating metrics for each class independently and then averaging the results, ensuring equal weight for all diagnostic categories regardless of their prevalence in the dataset. All performance metrics were calculated using scikit-learn and custom implementations.

\noindent To evaluate subgroup fairness, we additionally report protected-to-privileged ratios for AUROC, accuracy, true positive rate (TPR), false negative rate (FNR), false positive rate (FPR), positive predictive value (PPV), and negative predictive value (NPV). A ratio of 1.0 indicates parity between the protected and privileged groups. Ratios above or below 1.0 indicate subgroup disparity, although interpretation depends on the metric direction. For favorable metrics such as AUROC, ACC, TPR, PPV, and NPV, ratios below 1.0 indicate worse performance in the protected group, whereas ratios above 1.0 indicate better performance. For adverse error-based metrics such as FNR and FPR, ratios above 1.0 indicate higher error rates in the protected group and therefore a potential disadvantage. Because our intended use case is disease screening, FPR was emphasized in the main fairness analysis, while the remaining subgroup ratios were reported to provide a more complete assessment of disparity in discrimination, missed detection, false alarms, and predictive reliability across demographic groups.

\section{Implementation Details}
\noindent This study utilized a large-scale multi-imaging ophthalmic imaging dataset comprising paired 3D OCT and 2D en face OCT images. The primary pre-training corpus consisted of data from 32,765 patients collected at the University of Florida Health Eye Institute. Images were captured using standard clinical protocols on Heidelberg Spectralis OCT systems, ensuring pixel-to-voxel alignment. The 3D OCT volumes typically consisted of 19 or 25 B-scans with a resolution of $512 \times 512$ pixels, while the corresponding En face OCT images were $512 \times 512$ pixels. To ensure fair comparison, all experiments were conducted on the same hardware platform with NVIDIA B200 GPUs. All study procedures were conducted in accordance with the Declaration of Helsinki and approved by the institutional review boards of the participating institutions.


\section{Data availability}
This work uses retrospective UF Health data (IRB202300159) with a HIPAA Waiver of Authorization and safeguards to minimize privacy risk. The data were obtained from the University of Florida Integrated Data Repository (UF IDR) and contain protected health information. Therefore, the individual-level clinical data and imaging data used in this study are not publicly available and cannot be shared by the authors.

\section{Code availability}
The OphMAE framework used for this study can be found on GitHub. The analysis framework to evaluate all results, generate all plots, and perform all statistical analyses can be found in Supplementary Materials. All code uses Python v3.10, PyTorch v2.1.2, and Transformers v4.42.4. The code to create the dataset uses Python v3.10 and Pandas v2.1.3.

\section{Funding} 
This study is supported by the National Institutes of Health National Library of Medicine under Award Numbers R01LM014604 and R00LM014024, and the National Academy of Medicine under Award No. SCON-10001638. The content is solely the responsibility of the authors and does not necessarily represent the official views of the National Academy of Medicine. 

\section{Competing interests} 
The authors declare no competing interests.

\section{Online content}
Any methods, additional references, Nature Portfolio reporting summaries, source data, extended data, supplementary information, acknowledgments, peer review information, details of author contributions and competing interests, and statements of data and code availability are available at the link.

\newpage

\section{References}
\begin{spacing}{0.9}
\bibliographystyle{naturemag}
\bibliography{ref}
\end{spacing}

\newpage
\appendix

\section{UF Data Processing and Description}
\label{app:uf_dataset_description}

\begin{figure}[H]
  \centering
  \includegraphics[width=0.75\linewidth]{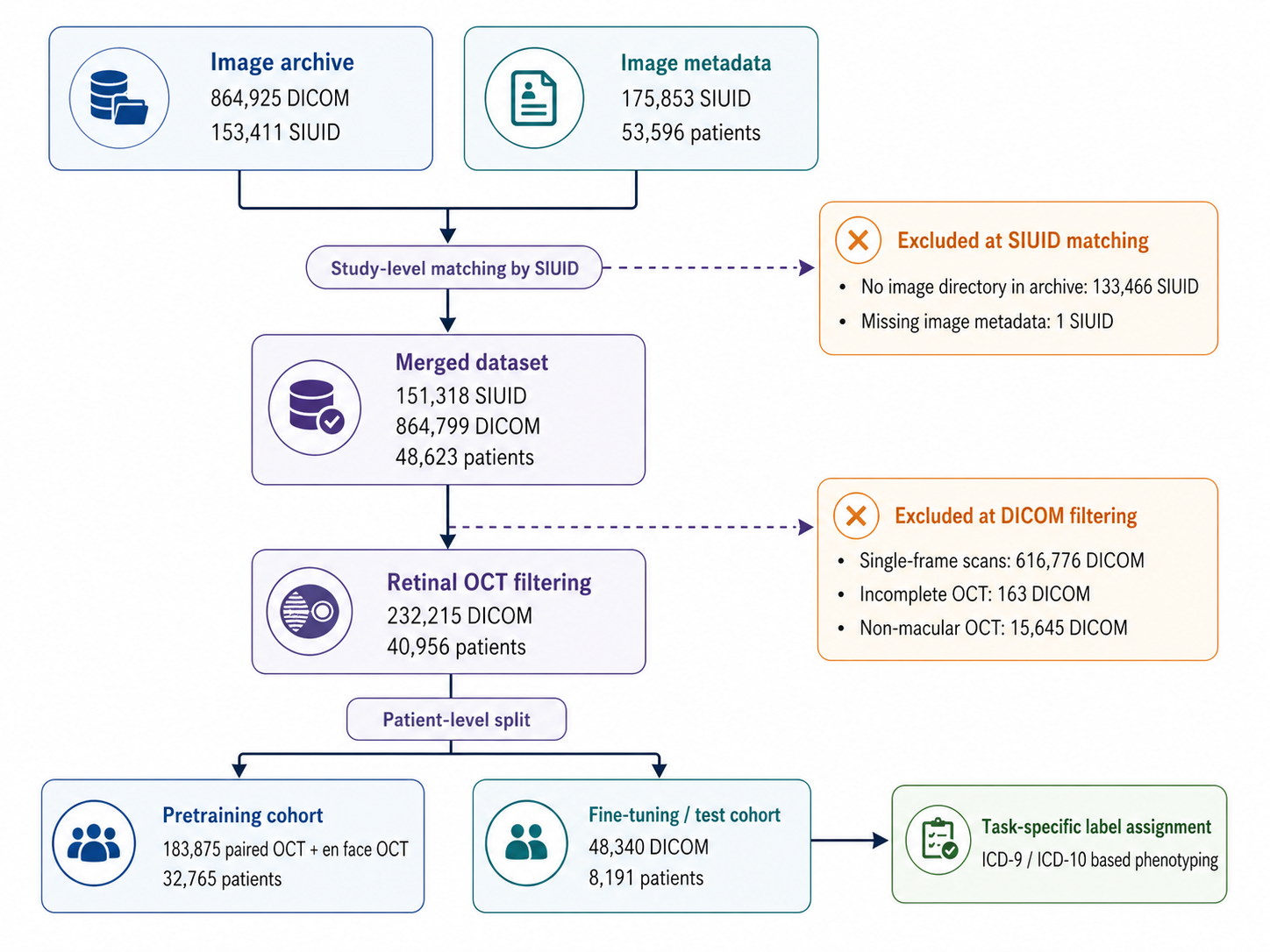}
  \caption{\textbf{Data curation and cohort construction pipeline for retinal OCT.} The pipeline summarizes the integration of raw retinal imaging archives and image metadata, followed by study-level matching, quality filtering, modality selection, patient-level dataset splitting, and downstream label assignment.}

  \label{fig:dataprocess}
\end{figure}

\noindent As shown in Figure \ref{fig:dataprocess}, the raw imaging archive contains (864,925 DICOM; 153,411 study instances (SIUID)) and image metadata (175,853 SIUID; 53,596 patients) and merges them by SIUID, yielding 151,318 SIUID with 864,799 DICOM from 48,623 patients. SIUIDs without a corresponding image directory (133,466 SIUID) or missing metadata (1) are excluded. We then filter DICOMs by validity and modality, removing studies with fewer than one frame (616,776 DICOM), completed/invalid entries (163 DICOM), and non-macular OCT scans (15,645 DICOM). Later, we divide the full dataset into pretrain and finetune datasets by patients. The pretrain dataset contains 32,765 patients with 183,875 DICOM files (a pair of OCT and En face OCT), and the finetune dataset has 8,191 patients with 48,340 DICOM files. Finally, we derive classification labels from the patient’s diagnosis history using ICD-9/ICD-10 codes in the finetune dataset. For control patients, we select the clean control patients who don’t have retinal-related disease and glaucoma ICD codes. The clean control dataset contains 1,148 patients and 3,099 DICOM. For each retinal diagnosis task, we choose the laterality (left, right, or both eyes) based on the ICD code and drop the latency to avoid labeling errors. For OCT images from control patients, we select the clean control patients who don’t have retinal-related disease and glaucoma ICD codes. Detailed pretrain/finetune/test data summaries and ICD code for each task are in the Supplement Table 1 and Supplement Table 2.

\noindent The Supplement Table 1 shows the UF dataset description of the full cohort, pre-trained, and finetune datasets. The full dataset contains 40,956 patients with 232,215 paired OCT images, and is then divided into a pretrain dataset (32,765 patients; 183,875 paired OCT) and a finetune dataset (8,191 patients; 48,340 paired OCT). The finetune dataset will be further filtered for clean controls and diseases OCT to avoid noise in the real-world dataset, and then divided into train:valid:test in a 4:1:5 ratio by patients. The detailed OCT and patient number for each eye diagnosis task and dataset split are shown in the supplement table 1.
\noindent Additionally, the supplement table 2 records the ICD codes we used for each task. For the ICD code without specified laterality, we will include both eyes. For control patients, we will exclude patients having disorders of the choroid and retina (H30-H36), Glaucoma (H40-H42), and disorders of the optic nerve and visual pathways (H46-H48).

\section{Complete Comparison of Diagnostic Performance across Diverse Pathologies}
\label{app:full_experiment}

\noindent As shown in Supplement Table 3, OphMAE achieved the highest AUROC and showed consistent overall performance across diverse retinal and systemic prediction settings. Beyond AUROC and F1 score, OphMAE was also reflected in class balance, agreement, and positive-case retrieval. In several representative binary tasks, OphMAE achieved the strongest or near-strongest average precision, Kappa, MCC, precision, and recall simultaneously. For AMD, OphMAE reached an average precision of 96.8\%, Kappa of 81.1\%, MCC of 81.1\%, precision of 90.4\%, and recall of 90.7\%. A similar pattern was observed for DME binary classification, where OphMAE achieved an average precision of 97.2\%, Kappa of 83.5\%, MCC of 83.6\%, precision of 91.7\%, and recall of 91.8\%. These consistent gains across complementary metrics support that the improvement of OphMAE was clinically meaningful rather than driven by AUROC alone.
\noindent The broader metric profile provides more information in more challenging multiclass settings. OphMAE achieved the strongest Kappa and MCC in multiclass glaucoma and DME, suggesting more stable label agreement under complex class boundaries. For systemic disease prediction, OphMAE again showed the most favorable balance across average precision, precision, recall, and MCC for both diabetes and DKD, despite the lower absolute performance of all models in these tasks.

\section{Complete Comparison of Subset Performance}
\label{app:sugroup}

\noindent Supplementary Table 4 provides a full subset-level comparison using accuracy, AUROC, F1 score, AUPRC, Kappa, MCC, precision, and recall. Across most tasks, performance declined only modestly when training data were reduced to 300 samples, but deteriorated more clearly in the extreme low-data setting of 100 samples, indicating that foundation-model benefits were largely preserved under moderate data reduction and became most challenged only under severe data scarcity.

\noindent Across the reduced-data experiments, OphMAE showed consistently competitive performance and maintained particular strength in CSR, DR binary, DKD, diabetes, glaucoma, cataract, and RNV. In these tasks, its AUROC advantages were generally accompanied by favorable AUPRC, MCC, and Kappa, supporting a more balanced predictive profile rather than an isolated gain in ranking performance. OphMAE also remained competitive for DME binary and CRVO/CRAO at 300 and 500 samples, further supporting its stability when fine-tuning data were limited.

\noindent The complete metric comparison also confirmed task-specific differences among foundation models. RETFound remained superior in several settings, such as AMD, ERM, and multiclass DME. Additionally, the complete result illustrates the need to interpret subset results beyond accuracy, particularly for imbalanced tasks. For example, models in the PD task have high accuracy but near-chance AUROC, AUPRC, F1, Kappa, and MCC. Overall, the full metric results reinforce that OphMAE is data efficient and stable under limited supervision.

\section{Complete Comparison of Subgroup Fairness Analysis}
\label{app:sugroup}

\noindent Supplement Table 5 summarizes subgroup fairness metrics for OphMAE and RETFound, including absolute differences in accuracy, AUROC, FNR, FPR, NPV, PPV, treatment equality, and TPR across age, sex, and race/ethnicity subgroups. Overall, AUROC and ACC ratios were generally closer to 1.0 than threshold-dependent metrics. This pattern suggests that global discrimination and overall accuracy were often relatively similar between subgroups, whereas clinically relevant differences were more likely to emerge in error-based and predictive value–based measures.

\noindent Age remained the subgroup factor with the largest and most variable disparity. Relative to patients aged 45–64 years, the group aged 75 years or older often showed larger deviations in TPR, FNR, PPV, and NPV, although the direction of these differences (ratio $>$ or $<$ 1.0) varied by task. In several tasks, including CRVO/CRAO, PVD, and Drusen binary classification, and CSR, older patients tended to show increased sensitivity, reflected by a TPR ratio greater than 1.0. By contrast, accuracy and AUROC ratios were usually shown to be lower than 1.0, indicating the weak performance in the older subgroup. By comparing the ratio of TPR, FPR, and FNR, the performance decreases mainly due to the model predicting more false positive cases.

\noindent Sex- and race/ethnicity-based comparisons showed a similar but weaker pattern. For both models, female-to-male and race/ethnicity subgroup ratios for AUROC and ACC were typically close to 1.0. Task-specific departures were still evident in TPR, FNR, and PPV, particularly for CRVO/CRAO, RNV, MH, and Cataract. For instance, OphMAE and RETFound have 2.44 and 1.56 PPV ratios comparing sex subgroups in the RNV task, and 1.95 and 1.40 PPV ratios in the CRVO/CRAO task. The results indicate that predictions are more reliable in the female subgroup. Moreover, the female subgroup demonstrates more advantages in AUROC and accuracy ratio in RETFound than OphMAE. In race/ethnicity-based comparisons, AMD, PVD, and MH tasks have a PPV ratio lower than 1.0, which indicates the reliability difference in the NHW subgroup.

\noindent The complete ratio-based analysis also refines the comparison between OphMAE and RETFound. Overall, OphMAE tended to be closer to parity than RETFound for AUROC, ACC, TPR, FNR, and especially NPV across many matched subgroup comparisons, whereas PPV parity was more mixed and remained task dependent. Neither model was uniformly closest to parity across all diseases, subgroup definitions, and metrics, underscoring that fairness conclusions depend on both the outcome and the metric selected.

\noindent Taken together, the complete subgroup analysis reinforces the interpretation of the main fairness results. The main-text emphasis on FPR is appropriate for a screening-oriented use case, but the additional ratio-based metrics show that subgroup fairness also depends on missed-detection rates, sensitivity, and predictive value balance. Across both models, AUROC and ACC alone would have understated these differences, because parity in overall discrimination often coexisted with substantial subgroup variation in TPR, FNR, PPV, or NPV. These findings support the use of multi-metric fairness evaluation when assessing ophthalmic foundation models for clinical deployment.

\section{Qualitative results of OphMAE}

\begin{figure}[H]
  \centering
  \includegraphics[width=0.8\linewidth]{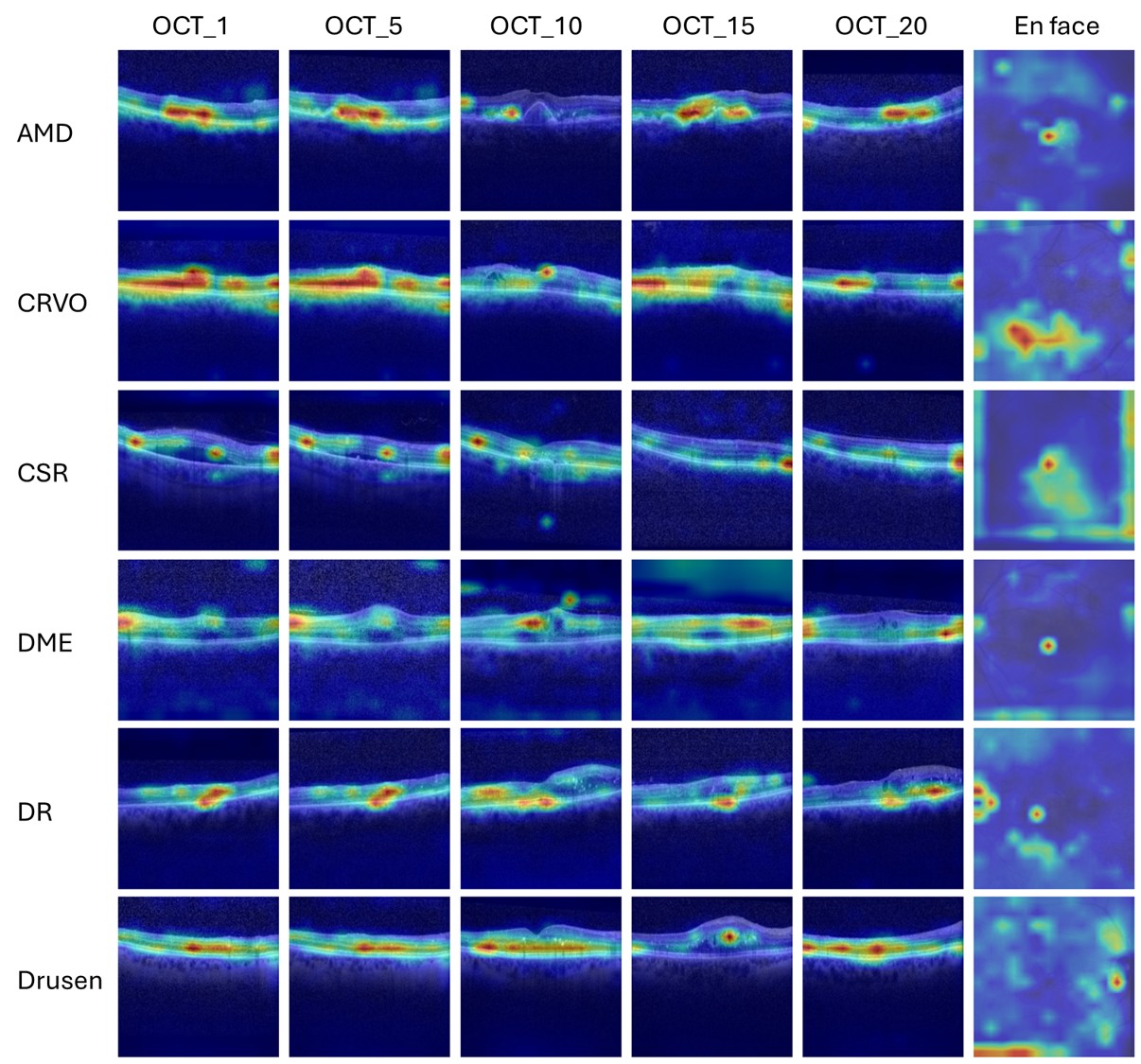}
  \caption{\textbf{Qualitative attribution maps of OphMAE for representative retinal diseases (set 1).} Representative heatmaps generated from the fine-tuned OphMAE model for AMD, CRVO, CSR, DME, DR, and drusen. Attribution maps are shown for both the 3D OCT volume and the paired 2D en face OCT image, with warmer colors indicating regions that contributed more strongly to the model prediction. For the volumetric branch, heatmaps are overlaid on representative OCT slices to illustrate the spatial distribution of model attention across depth.}
  \label{fig:xai_case1}
\end{figure}

\begin{figure}[H]
  \centering
  \includegraphics[width=0.8\linewidth]{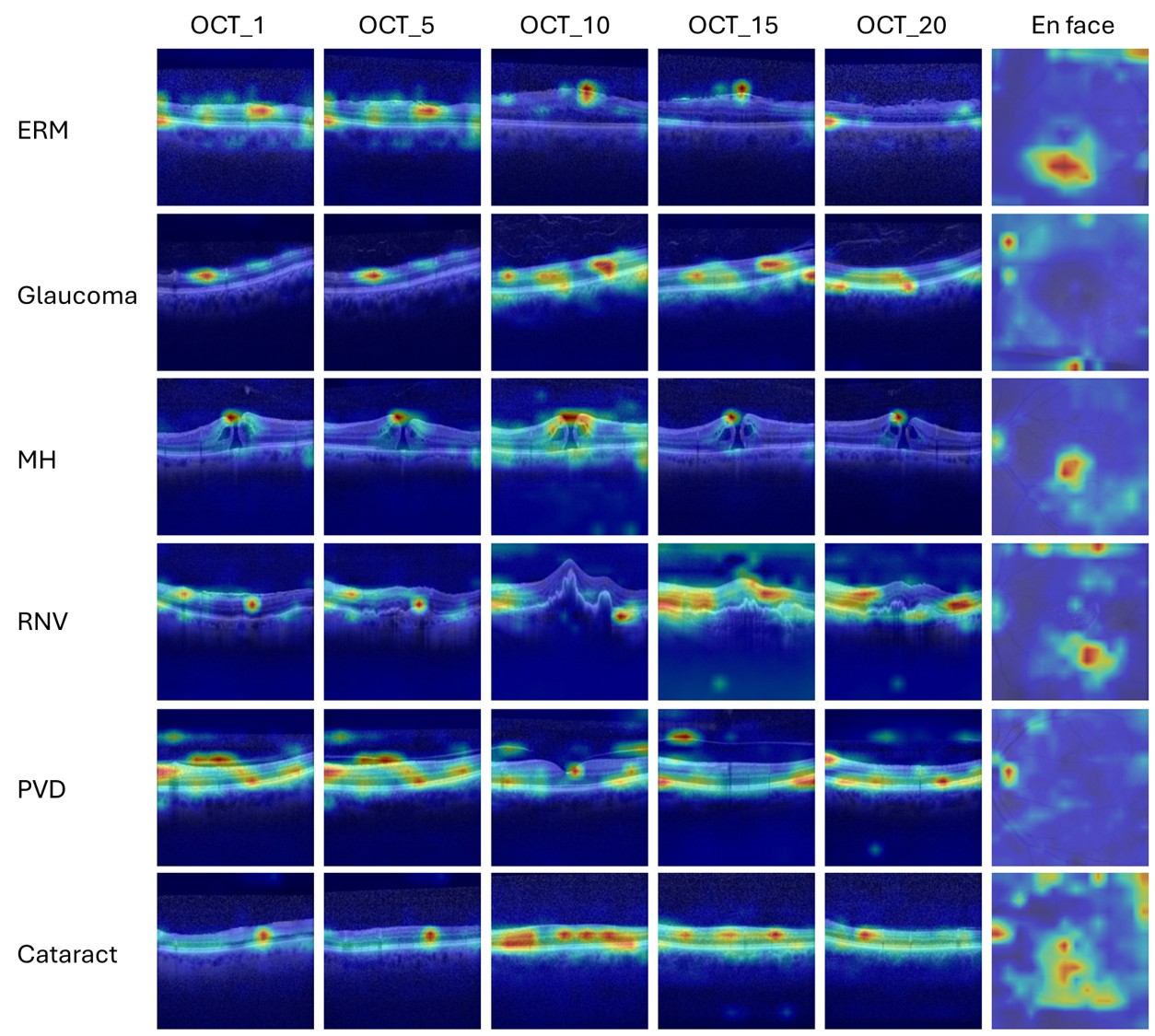}
  \caption{\textbf{Qualitative attribution maps of OphMAE for representative retinal diseases (set 2).} Representative heatmaps generated from the fine-tuned OphMAE model for AMD, CRVO, CSR, DME, DR, and drusen. Attribution maps are shown for both the 3D OCT volume and the paired 2D en face OCT image, with warmer colors indicating regions that contributed more strongly to the model prediction. For the volumetric branch, heatmaps are overlaid on representative OCT slices to illustrate the spatial distribution of model attention across depth.}
  \label{fig:xai_case2}
\end{figure}

To better understand the internal focus of OphMAE, we performed a qualitative heatmap analysis of the fine-tuned model using Transformer Explainability. Figures \ref{fig:xai_case1} and \ref{fig:xai_case2} show attribution maps for both the 3D OCT volume and the 2D en face OCT image, with warmer colors indicating regions that contributed more strongly to the final prediction. For visualization of the volumetric branch, we overlaid the heatmaps on representative OCT slices at depths 1, 5, 10, 15, and 20 to illustrate how model attention was distributed across the OCT volume.

\noindent Overall, the attribution maps showed that OphMAE consistently focused on clinically relevant regions across multiple disease tasks. In AMD, the model highlighted areas corresponding to pathologic retinal changes; in CSR, it emphasized regions associated with subretinal fluid and retinal detachment; in DR and DME, the highlighted regions were often located around areas of fluid accumulation; in drusen, the model attended to hyper-reflective abnormalities; and in ERM, the heatmaps concentrated near the membrane and the distorted retinal surface. These qualitative findings suggest that OphMAE learned disease-relevant retinal representations rather than relying on diffuse or anatomically implausible image features.

\noindent An additional observation was that the most highly attributed regions were often located adjacent to, rather than directly on, the lesion itself. For example, in MH, the strongest attribution was not centered on the hole alone, but on the disrupted retinal margins surrounding the lesion. Similarly, highly attributed regions frequently appeared at interfaces between fluid spaces and retinal layers. This pattern may reflect the patch-based tokenization used by transformer models, in which the local context is aggregated over neighboring pixels. These heatmaps suggest that OphMAE bases its predictions on morphologically meaningful structures and local lesion context.
 
\end{document}